\documentclass[sigconf]{acmart}

\usepackage{amsmath}
\usepackage{multirow}
\usepackage{makecell}
\usepackage{amssymb}
\usepackage{graphicx}
\usepackage{dsfont}
\usepackage{amssymb}
\usepackage{color, colortbl}
\usepackage{pifont}
\usepackage{hhline}
\usepackage{amsmath}
\usepackage{subfigure}
\usepackage{enumerate}

\AtBeginDocument{%
  \providecommand\BibTeX{{%
    \normalfont B\kern-0.5em{\scshape i\kern-0.25em b}\kern-0.8em\TeX}}}

\setcopyright{acmcopyright}
\copyrightyear{2020}
\acmYear{2020}
\setcopyright{acmcopyright}\acmConference[MM '20]{Proceedings of the 28th ACM International Conference on Multimedia}{October 12--16, 2020}{Seattle, WA, USA}
\acmBooktitle{Proceedings of the 28th ACM International Conference on Multimedia (MM '20), October 12--16, 2020, Seattle, WA, USA}
\acmPrice{15.00}
\acmDOI{10.1145/3394171.3413776}
\acmISBN{978-1-4503-7988-5/20/10}

\begin{document}

\fancyhead{}

\title{Emotion-Based End-to-End Matching Between Image and Music in Valence-Arousal Space}

\author{Sicheng Zhao}
\authornote{Corresponding author.}
\authornote{Equal contribution.}
\email{schzhao@gmail.com}
\affiliation{%
  \institution{University of California, Berkeley}
}

\author{Yaxian Li}
\authornotemark[2]
\email{liyaxian@ruc.edu.cn}
\affiliation{%
  \institution{Renmin University of China}
}

\author{Xingxu Yao}
\authornotemark[2]
\email{yxx\_hbgd@163.com}
\affiliation{%
  \institution{Nankai University, Didi Chuxing}
}

\author{Weizhi Nie}
\email{weizhinie@tju.edu.cn}
\affiliation{%
  \institution{Tianjin University}
}

\author{Pengfei Xu}
\email{xupengfeipf@didiglobal.com}
\affiliation{%
  \institution{Didi Chuxing}
}

\author{Jufeng Yang}
\email{yangjufeng@nankai.edu.cn}
\affiliation{%
  \institution{Nankai University}
}

\author{Kurt Keutzer}
\email{keutzer@berkeley.edu}
\affiliation{%
  \institution{University of California, Berkeley}
}


\begin{abstract}
Both images and music can convey rich semantics and are widely used to induce specific emotions. Matching images and music with similar emotions might help to make emotion perceptions more vivid and stronger. Existing emotion-based image and music matching methods either employ limited categorical emotion states which cannot well reflect the complexity and subtlety of emotions, or train the matching model using an impractical multi-stage pipeline. In this paper, we study end-to-end matching between image and music based on emotions in the continuous valence-arousal (VA) space. First, we construct a large-scale dataset, termed Image-Music-Emotion-Matching-Net (IMEMNet), with over 140K image-music pairs. Second, we propose cross-modal deep continuous metric learning (CDCML) to learn a shared latent embedding space which preserves the cross-modal similarity relationship in the continuous matching space. Finally, we refine the embedding space by further preserving the single-modal emotion relationship in the VA spaces of both images and music. The metric learning in the embedding space and task regression in the label space are jointly optimized for both cross-modal matching and single-modal VA prediction. The extensive experiments conducted on IMEMNet demonstrate the superiority of CDCML for emotion-based image and music matching as compared to the state-of-the-art approaches.
\end{abstract}



\maketitle

\begin{figure}[!t]
	\centering
	\includegraphics[width=0.95\linewidth]{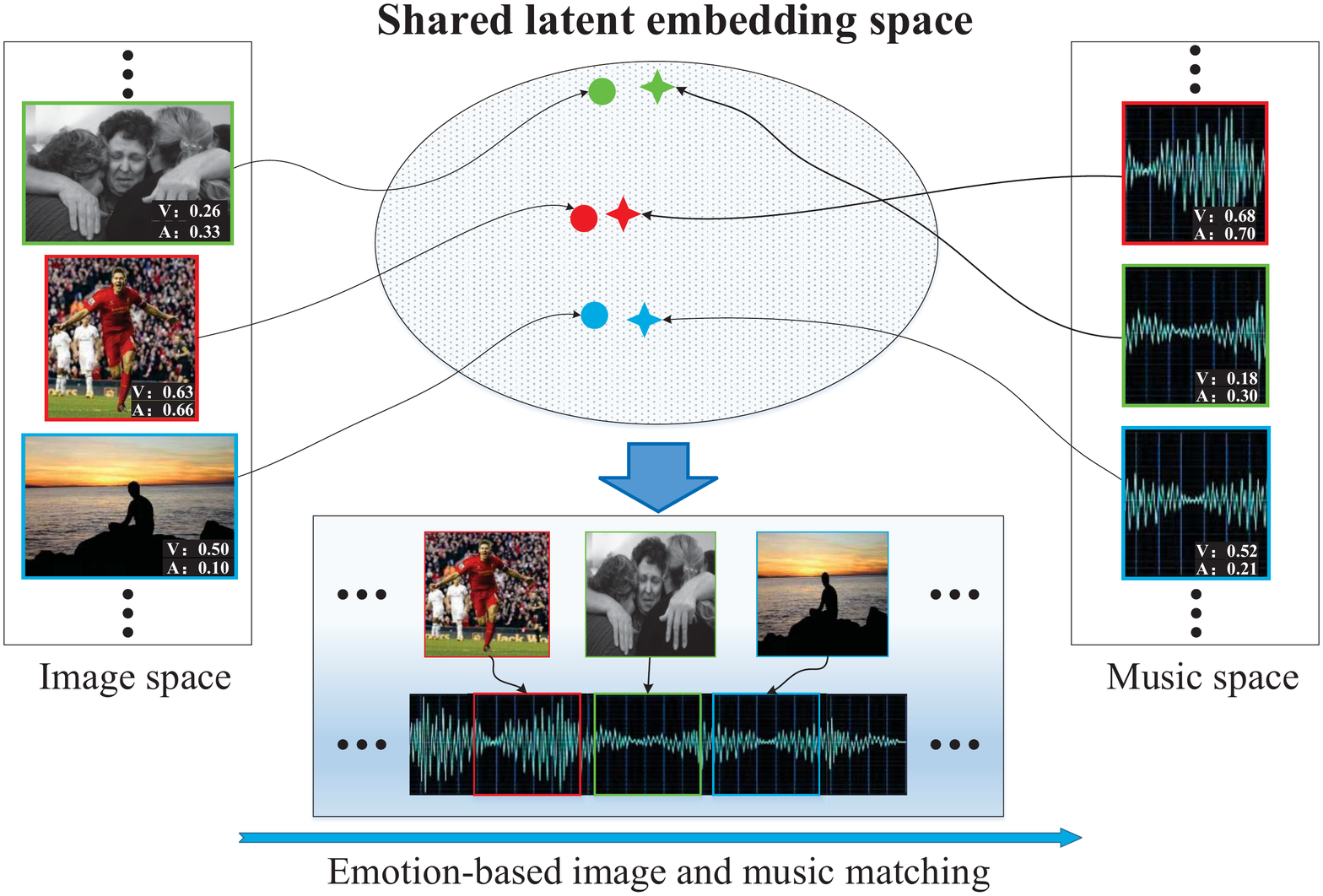}\\
	\caption{The basic idea of our image and music matching based on continuous emotions. Both images and music are projected into the same shared latent embedding space, which is learned by preserving both the cross-modal and single-modal emotion relationships.}
	\label{fig:motivation}
\end{figure}

\section{Introduction}

Humans are emotional animals. Famous artists remain immortal because the artworks they create such as paintings, music, and literary works can express unique insights of life and cause emotional resonance to the audience~\cite{gaut2007art,zhao2014exploring}. The wide popularity of mobile devices and social networks enables everyone to become an ``artist''. Humans habitually use images, audios, and videos
together with text in social networks to express their opinions and share their emotions~\cite{zhao2018predicting,zhao2020end}. Affective analysis of the huge volume of multimedia data can help to understand humans' behaviors and preferences and thus plays an important role in many practical applications~\cite{zhao2019pdanet,she2020wscnet},
such as opinion mining~\cite{truong2017visual,ye2019visual,hassan2019sentiment}, business intelligence~\cite{holbrook1984role,poels2006capture,hosany2010measuring,pan2014travel,toyama2016categorization,hosany2013patterns}, psychological health~\cite{lin2014psychological,bao2014thupis,guntuku2019twitter}, and entertainment assistant~\cite{tan2008entertainment,xing2015emotion,ahmad2012emotion,yang2019human}.

Recently, extensive efforts have been dedicated to recognizing the emotions in single-modality, such as text~\cite{giachanou2016like,zhang2018deep}, image~\cite{joshi2011aesthetics,zhao2018affective}, speech~\cite{el2011survey}, and music~\cite{yang2012machine}. There are also increasing efforts on fusing information from multiple modalities~\cite{wang2015video,soleymani2017survey,poria2017review,zhao2019affective}. Since different modalities can provide complementary ability for emotion recognition, these multi-modal based methods usually achieve better performance. The main goal is to bridge the affective gap by extracting discriminative features and designing effective learning or fusing strategies~\cite{zhao2019affective}.

Compared with multi-modal emotion recognition, relatively less efforts have been made to understand the emotion-centric correlation between different modalities (\textit{e.g.} image and music studied in this paper). Such emotion-based matching is essential for various applications~\cite{verma2019learning}, such as affective cross-modal retrieval, emotion-based multimedia slideshow, and emotion-aware recommendation systems. The early emotion-based matching methods mainly employ a shallow pipeline~\cite{chen2008emotion,su2011photosense,xiang2012synaesthetic,sasaki2013affective,zhao2014emotion,lee2016system}, \textit{i.e.} extracting hand-crafted features and training matching classifiers (or training emotion classifiers for both modalities and then learning matching similarities). The differences lie in the extracted features, employed emotion representations, learned classifiers, and similarity metrics. Until very recently, some methods~\cite{
verma2019learning,xing2019image} train a matching model end-to-end with specific emotion categories by concatenating the extracted visual and audio features and feeding them into a few fully-connected (FC) layers.

%

However, there are some limitations of existing emotion-based image and music matching methods. First, many employ limited categorical states to represent emotion. As recent psychological theories show, emotion categories in real-world are actually diverse and fine-grained~\cite{cowen2017self,zhan2019zero}. Therefore, such coarse-grained representation cannot well reflect the complexity and subtlety of emotions. Second, most models are trained in a multi-stage pipeline, which are impractical. Third, they do not consider how the emotional content is shared in a latent embedding space across different modalities or the learned space cannot guarantee to preserve the relationship in the label space. Finally, they do not release the datasets, which make it difficult to compare with these methods.

To address the above-mentioned problems, we propose to match image and music based on continuous valence-arousal emotions in an end-to-end manner, as shown in Figure~\ref{fig:motivation}. We construct Image-Music-Emotion-Matching-Net (IMEMNet), a large-scale dataset for evaluation with over 140K image-music pairs. We select DEAM~\cite{alajanki2016benchmarking} as the music corpora, and combine IAPS~\cite{lang1997international}, NAPS~\cite{marchewka2014nencki}, and EMOTIC~\cite{kosti2017emotion} as the image corpora. To project the image and music modalities into the same shared latent embedding space, we propose cross-modal deep continuous metric learning (CDCML), which consists of three components. Cross-modal similarity metric learning enforces the distance ratios in the cross-modal matching space to be preserved in
the learned embedding space. Single-modal emotion metric learning further refines the embedding space by preserving the distance ratios in the VA space of both images and music. Embedded multi-task regression learns desired regression models based on the embeddings for multi-task continuous predictions: cross-modal similarities and single-modal VA values.


In summary, the contributions of this paper are threefold:

(1) We are the first to study image and music matching based on continuous valence-arousal emotions in an end-to-end manner.

(2) We propose a novel metric learning method, CDCML, to match image and music based on emotions by learning a shared latent embedding space. The joint optimization of metric learning in the embedding space and task regression in the label space enables CDCML to simultaneously predict cross-modal matching similarities and single-modal VA values.

(3) We construct a new large-scale dataset, termed IMEMNet, for continuous emotion-based image and music matching. Extensive experimental results on IMEMNet demonstrate that the proposed CDCML method outperforms the state-of-the-art methods by a large margin for emotion-based image and music matching.


\section{Related Work}
\label{sec:relatedWork}

\noindent\textbf{Emotion Representation Models.}
In psychology, emotion is often measured by two kinds of representation models: categorical emotion states (CES) and dimensional emotion space (DES).
CES models aim to classify emotions into several discrete categories, which are easy to understand for non-professionals.
The simplest CES model is sentiment polarity, \textit{i.e.} positive and negative.
More emotion categories are proposed based on psychological theories, such as Mikel's eight emotions~\cite{mikels2005emotional} and Ekman's six emotions~\cite{ekman1992argument}.

To more accurately model the complexity and subtlety of emotions, an increasing number of psychology studies tend to represent emotions using DES models in a 2D, 3D, or higher dimensional Cartesian space.
One most popular DES model is valence-arousal-dominance (VAD)~\cite{schlosberg1954three}, where valence denotes the degree of pleasantness ranging from positive and negative, arousal shows the intensity of emotion ranging from excited to calm, and dominance represents the level of control ranging from controlled to in control.
Due to the difficulty in predicting dominance, many studies represent emotions in VA space~\cite{hanjalic2006extracting,kim2018building,zhao2017continuous}.
In this paper, we develop an end-to-end framework for cross-modal matching between image and music based on continuous emotions in VA space.

\noindent\textbf{Image Emotion Recognition.}
The studies for image emotion recognition emerge in large numbers recently, which originates from the research in psychology to explore the relation between visual stimuli and emotion~\cite{lang1997international, mikels2005emotional}.
In the earlier years, many types of hand-crafted representations~\cite{machajdik2010affective, zhao2014exploring} are designed to bridge affective gap between low-level features and abstract emotions, such as adjective noun pairs~\cite{borth2013sentibank, chen2014deepsentibank} and high-level concepts~\cite{ali2017high}.
With the success of the convolutional neural networks (CNNs) on different multimedia tasks, current researchers mainly design CNN-based algorithms~\cite{she2020wscnet,you2015robust,you2016building,yang2018visual,yang2018retrieving,zhao2018emotiongan,yao2019attention,zhao2019cycleemotiongan}.
In CES model, apart from traditional dominant emotion classification, label distribution learning~\cite{yang2017joint,zhao2017approximating,zhao2017learning} is introduced to tackle the ambiguity of image emotion by describing each category with a concrete probability.
Using DES model, \citeauthor{kim2018building}~\cite{kim2018building} developed an emotion-based network that combines low-level features, object, and background information to predict emotion values in VA space.
In~\cite{zhao2019pdanet}, polarity-consistent regression loss is designed to take emotion's polarity into account for VAD prediction.
Differently, our method not only penalizes the VA predictions, but also considers the feature distance in an embedding space based on the emotion similarity in VA space.

\noindent\textbf{Music Emotion Recognition.}
Over the years, various methods have emerged to characterize and quantify the emotions associated with music. The early music emotion recognition methods mainly implement traditional machine learning algorithms with hand-crafted acoustic features as input~\cite{wang2012acoustic,deng2015emotional,patra2013unsupervised,xianyu2016svr,wu2014music}, the validity and generality of which cannot be guaranteed~\cite{dong2019bidirectional}. Since these methods require careful design and data preprocessing based on extensive prior knowledge, recent emphasis has been shifted to automatically extracting features from the original data. Representative methods include CNNs~\cite{mao2014learning,han2016deep}, recurrent neural networks (RNNs) especially long short-term memory (LSTM)~\cite{cai2016maxout,chen2017multimodal}, and the combination of CNN and RNN~\cite{lim2016speech,adavanne2017stacked,dong2019bidirectional}. Similar to image emotion recognition, we also preserve the music emotion similarity when learning the embedding space.

\noindent\textbf{Emotion-Based Image and Music Matching.}
\citeauthor{chen2008emotion}~\cite{chen2008emotion} proposed to visualize music using photos based on their emotion categories. They separately extracted hand-crafted features, learned emotion classifiers, and composited images and music based on the predicted emotions. Many methods follow this pipeline~\cite{su2011photosense,xiang2012synaesthetic,sasaki2013affective,zhao2014emotion,lee2016system}. They (1) extracted more discriminative emotion features, such as low-level color~\cite{chen2008emotion,su2011photosense,xiang2012synaesthetic,sasaki2013affective,lee2016system} and mid-level principles-of-art~\cite{zhao2014emotion} for image; (2) employed different emotion representation models, from categorical states~\cite{chen2008emotion,su2011photosense,xiang2012synaesthetic,lee2016system} to dimensional space~\cite{sasaki2013affective,zhao2014emotion}; (3) correspondingly learned different classifiers, from Support Vector Machine~\cite{chen2008emotion}, Naive Bayes, and Decision Tree~\cite{su2011photosense} to Support Vector Regression~\cite{zhao2014emotion}; and (4) used different composition strategies to match image and music, from emotion category comparison~\cite{chen2008emotion,su2011photosense,xiang2012synaesthetic,lee2016system} to Euclidean distance~\cite{sasaki2013affective,zhao2014emotion}.



The most relevant methods to ours are~\cite{verma2019learning,xing2019image}. \citeauthor{verma2019learning}~\cite{verma2019learning} proposed to learn affective correspondence between image and music based on sentiment polarity (positive, negative, and neutral). The images and music are projected into a common representation space and a binary classification task is performed to predict the affective correspondence by a few fully-connected (FC) layers. \citeauthor{xing2019image}~\cite{xing2019image} studied a similar task but the dataset is collected using Chinese folk images and music, which are annotated using Hevner Emotion Ring model with eight emotion categories. They also investigated the emotion similarity comparison approaches between Pearson correlation coefficient and Euclidean distance.

Differently, we propose to match image and music based on continuous emotions to better reflect the complexity and subtlety of emotions. Further, the projected latent embedding space preserves the relationship in the cross-modal similarity space and in the single-modal emotion space.


\noindent\textbf{Deep Metric Learning.}
Deep metric learning has been widely utilized to measure the similarity or distance between different samples.
As the standard loss functions, contrastive loss~\cite{chopra2005learning} and triplet loss~\cite{schroff2015facenet} are milestones of deep metric learning and are widely employed in subsequent work.
The contrastive loss minimizes the distance of samples from the same classes, and separates the samples of different classes away with a fixed margin.
The triplet loss introduces three types of samples, named anchor, positive, and negative samples.
Specifically, the loss enforces the distance between the anchor and the negative to be larger than that between the anchor and the positive.
To improve the efficiency of metric learning, \citeauthor{oh2016deep}~\cite{oh2016deep} utilized a matrix comprising pairwise distance of the mini-batch to design a loss, in which a lifted embedding structure is formed by all samples.
Simultaneously, $n$-pair loss aims to learn the embeddings for $(n+1)$-tuple, including an anchor, a positive, and $n-2$ negative examples.

In the field of cross-modal matching or retrieval across multimedia data such as image, text, and audio, deep metric learning is broadly used to transform the features of each modality into a common embedding space~\cite{peng2016cross, zhen2019deep}.
In~\cite{liong2016deep}, \citeauthor{liong2016deep} designed a unified architecture including two parallel neural networks, in which the intra-class variation is minimized and the inter-class variation is enlarged, and the difference of each sample pair from two modalities of the same class is minimized, respectively.
\citeauthor{kang2017learning}~\cite{kang2017learning} integrated the center loss and softmax cross-entropy loss to learn an embedding space that has a semantic meaning for both image and text for cross-modal retrieval.

As emotions in VA space are continuous values, the binary supervision that indicates whether a pair of data belong to the same class cannot describe the similarity. Inspired by log-ratio loss~\cite{kim2019deep}, we propose cross-modal deep continuous metric learning to measure the degree of continuous cross-modal similarity.


\section{The IMEMNET Dataset}
\label{sec:Dataset}
In this section, we introduce the IMEMNET dataset\footnote{The IMEMNet dataset is released at: \url{https://github.com/linkAmy/IMEMNet}.} on continuous emotion-based image and music matching, including image and music data selection and image-music matching.


\begin{table}[!t]
\caption{Statistics of the IMEMNet dataset, where `\#' denotes the corresponding number (the same below).}
\begin{tabular}{lrrrr}
\toprule
         & Training& Validation  & Testing & Total \\ \hline
\#Songs  & 1,442  & 90  & 270   & 1,802  \\
\#Song clips  & 28,835  & 1,759  & 5,223   & 35,817  \\
\#Images & 20,496 & 1,281 & 3,843 & 25,620 \\
\#Paires & 109,525 & 8,795 & 26,115 & 144,435 \\
\bottomrule
\end{tabular}
\label{tab:datasplit}
\end{table}

\begin{table*}[]
\caption{Comparison of our released IMEMNet dataset with others, where the values in the parentheses of the second column are the number of emotion categories or detailed emotion space, `ED', `AED', and `PCC' are abbreviations for Euclidean distance, Aesthetic energy distance, and Pearson correlation coefficient, respectively.}
\begin{tabular}{l|lrrrrr|c}
\toprule
Reference                                          & Emotion label & \#Images & \#Music & Clip   length & \#Pairs           & Matching                    & Released \\ \hline
~\cite{chen2008emotion}       & CES (8)        & 368      & -      & 5s                  & -                  & Self-defined       & No      \\
~\cite{su2011photosense}   & DES (VA)       & 3,000    & 1000    & 30s                 & - & ED          & No      \\
~\cite{xiang2012synaesthetic} & CES (3)        & 233      & 16      & Unfixed     & -                  & AED & No      \\
~\cite{sasaki2013affective}   & DES (VA)       & 1,182    & 315     & Unfixed                & -            & ED          & No      \\
~\cite{zhao2014emotion}       & DES (VA)       & 1,182    & 240     & 15s                 & -                  & ED         & No      \\
~\cite{lee2016system}         & DES (VA)       & 57       & 273     & 20s                 & -            & ED          & No      \\
~\cite{verma2019learning}     & CES (3)        & 85,000   & 3,812    & 60s                 & -                  & 0/1                         & Yes     \\
~\cite{xing2019image}   & CES (8)        & 500      & 500     & 30s                 & 250,000            & ED \& PCC   & No      \\ \hline
\textbf{Ours}                                       & DES (VA)       & 25,620   & 1,802   & 2s                  & 144,435            & ED         & Yes     \\
\bottomrule
\end{tabular}
\label{tab:datasetComparison}
\end{table*}

\subsection{Image and Music Data Selection}

We combine IAPS~\cite{lang1997international}, NAPS~\cite{marchewka2014nencki}, and EMOTIC~\cite{kosti2017emotion} with continuous VA labels as the image corpora. IAPS is an emotion evoking image set in psychology with 1,182 documentary-style natural color images. Each image is annotated with a 9-point VAD ratingby about 100 college students. NAPS consists of 1,356 realistic, high-quality photographs rated by 204 mostly European participants in a 9-point bipolar semantic sliding scale on VA and approach-avoidance dimensions. EMOTIC is a dataset with 23,082 images containing people in non-controlled environments. The images were annotated by Amazon Mechanical Turk (AMT) workers with continuous 10-scale VAD dimensions.

We select DEAM~\cite{alajanki2016benchmarking} as the music corpora. DEAM consists of 1,802 excerpts and full songs annotated with VA values (from -1 to +1) both continuously (per-second) and over the whole song. Considering the stability of the annotations, each song is annotated from the 15th second. The frequency of all songs is 44100Hz. Most of the songs (1,723 in total) are 45 seconds long, with the rest varying in length, reaching a maximum of more than 600 seconds.


Since the image and music data is labeled in different scales, we normalize the VA values into [0,1] respectively based on the minimum and range. After normalization, we randomly split both image and music data into 80\% for training, 5\% for validation, and 15\% for testing, as shown in Table~\ref{tab:datasplit}.

\subsection{Image-Music Matching}
\label{imm}
To match the images and music clips, we calculate the Euclidean distance between their VA ground truth labels, and then obtain the similarity as follows:
\begin{equation}
S(I_i, M_j)=\exp \bigg(-\frac{d\big(y^{I_i},y^{M_j}\big)}{\sigma_{n}^{m}}\bigg),
i = 1,\cdots,n,\ j = 1,\cdots,m,
\label{eq:eu2sim}
\end{equation}
where $d$ stands for the Euclidean distance, $y^{I_i}$ and $y^{M_j}$ are the VA labels of image $I_{i}$ and music clip $M_{j}$, $n$ and $m$ are the numbers of images and music clips, respectively.
$\sigma_{n}^{m}$ is set as the average Euclidean distance between all images and music clips. The degree of similarity is then set as the emotion matching label for corresponding image and music clip.

It is worth noting that all possible matching pairs is $m\times n$. For our images and music clips, the number of matching pairs will reach hundreds of millions. In order to avoid the explosion of the dataset scale, for each music clip, we select 50 images. Among them, 30 are randomly selected from the image dataset, and the remaining 20 are composed of 10 with highest matching score and 10 with the lowest. Finally, we randomly sample 10\% of the pairs to constitute the IMEMNet dataset. Please note that the images and music clips of the training set, verification set, and test set do not intersect. They are constructed independently. The statistics of the IMEMNet dataset is summarized in Table~\ref{tab:datasplit}, and the comparison of IMEMNet with existing datasets are compared in Table~\ref{tab:datasetComparison}.

\section{Problem Definition}
\label{sec:Definition}

In this paper, we study the problem of continuous emotion-based matching between images $\mathcal{I}$ and music $\mathcal{M}$, where $\mathcal{I} = \left\{I_i\right\}^n_{i=1}$ and $\mathcal{M} = \left\{M_i\right\}^m_{i=1}$.
%
%
On one hand, we aim to predict the degree of similarity between images and music clips; on the other hand, we also aim to predict the concrete VA values for each modality.	
Given the dataset consisting of $N$ image-music pairs $\mathcal{P}=\left\{(I_i, M_i)\right\}^{N}_{i=1}$ and their ground truth on the degree of similarity $\mathcal{S}=\left\{S(I_i, M_i)\right\}^{N}_{i=1}$, we build a branch $F_1: \mathcal{P}\rightarrow \mathcal{S}$ to learn the similarity for the input sample pairs.
Meanwhile, the single image or music clip has its own ground truth VA values.
Specifically, we use $\mathcal{Y^I} = \left\{y^{I_i} \right\}^{n}_{i=1}$ and $\mathcal{Y^M} = \left\{y^{M_j} \right\}^{m}_{j=1}$ to represent the emotion labels of images and music clips, respectively, where $y^{I_i}=(v^{I_i}, a^{I_i})$ represent the valence and arousal of the $i^{th}$ image and $y^{M_j}=(v^{M_j}, a^{M_j})$ represent the valence and arousal of the $j^{th}$ music clip.
Therefore, our another objective is to learn a common branch to learn the mapping $F_2: \mathcal{I}\rightarrow \mathcal{Y^I} $ and $\mathcal{M}\rightarrow \mathcal{Y^M} $.

\begin{figure*}[t]
	\centering
	\includegraphics[width=0.95\linewidth]{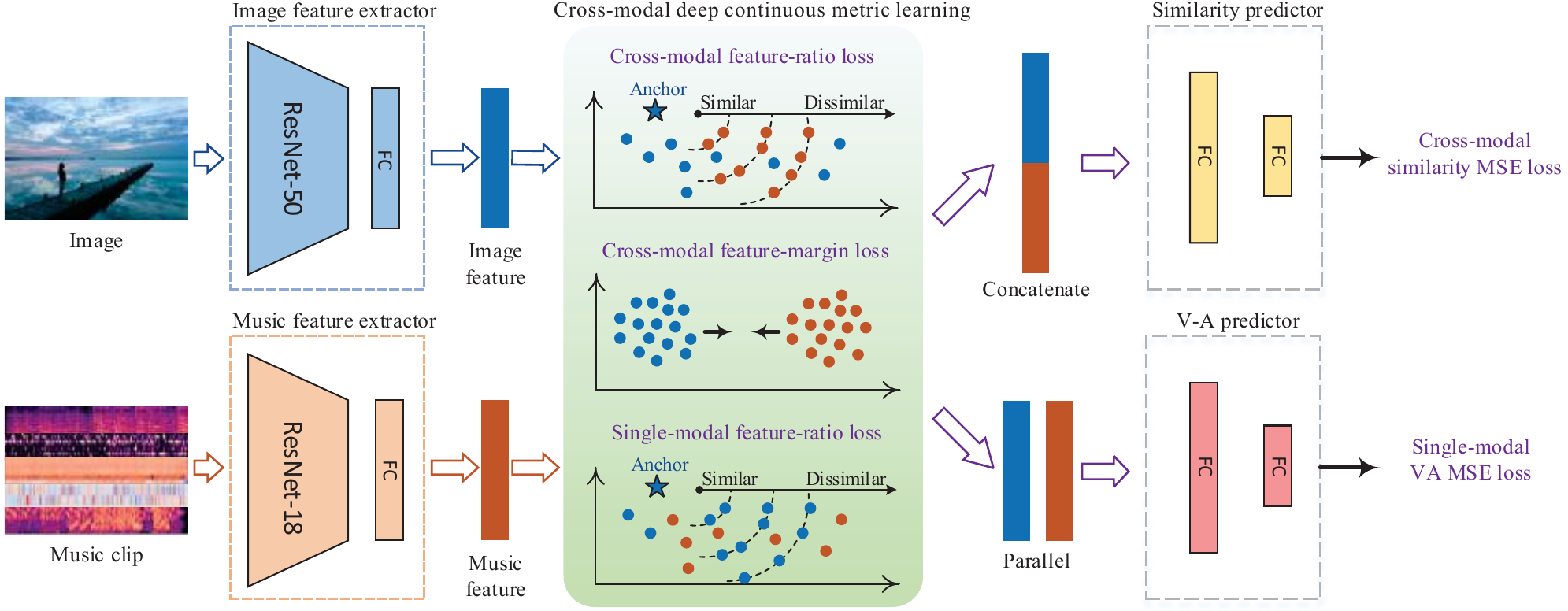}\\
	\vspace{2mm}
	\caption{Framework of the proposed CDCML for continuous emotion-based image and music matching. The blue and orange circles represent image samples and music clips. `FC' represents fully-connected layers. For simplicity, we omit the cross-modal feature-ratio loss and single-modal feature-ratio loss using music clips as anchors.}
	\label{fig:pip}
\end{figure*}

\section{Cross-modal Deep Continuous Metric Learning}
\label{sec:Approach}
In this section, we introduce the detailed cross-modal deep continuous metric learning (CDCML). The framework is shown in Figure~\ref{fig:pip}. First, we feed images and music clips into two parallel feature extractors, which take ResNet-50 and ResNet-18~\cite{he2016deep} as backbones, respectively. Second, we employ cross-modal similarity metric learning and single-modal emotion metric learning to learn a shared latent embedding space by optimizing various metric losses. The cross-modal similarity and single-modal emotion relationships are well preserved in the embedding space. Finally, we jointly learn a similarity predictor for image and music matching and a VA predictor for continuous emotion regression.

%
%

\subsection{Cross-modal Similarity Metric Learning}
\label{ssec:Cross-modal}

In the shared feature space, we refine the feature distributions to reflect the similarity relationships and to minimize the gap between image and music modalities.

%

\subsubsection{Cross-modal feature-ratio Loss}
The popular metric learning methods usually enlarge the inter-class variation and minimize the intra-class variation by modulating the Euclidean distance between features.
However, there is no specific class in VA space, where emotion label is continuous, so the widely-used metric losses cannot be directly applied in our tasks.
Inspired by~\cite{kim2019deep}, we propose a cross-modal feature-ratio loss $L_{CFR}$ to accurately optimize the distance of multi-modal embeddings based on their emotion similarity:
%
%
\begin{equation}
	\begin{aligned}
		\label{met}
	L_{CFR} &= \sum_{i=1}^N \left\{ \log \frac{D(f^{I_i}, f^{M_i})}{D( f^{I_i},f^{M_j}) }-\log \frac{S(I_i, M_i)}{S(I_i, M_j) } \right\}^2 \\
	&+ \sum_{i=1}^N \left\{ \log \frac{D(f^{M_i}, f^{I_i})}{D( f^{M_i},f^{I_j}) }-\log \frac{S(M_i, I_i)}{S(M_i, I_j) } \right\}^2,\\
	\end{aligned}%
\end{equation}
where $D(\cdot)$ means the squared Euclidean distance, $S(\cdot)$ denotes the similarity between image and music clip as defined in Eq.~(\ref{eq:eu2sim}), and $i \neq j$.
%
%
In the first item, image $I_i$ is treated as the anchor, while the music clip $M_i$ serves as the anchor in the second item.
For the anchor of one modality (\textit{e.g.} $I_i$), we randomly choose two examples from another modality (\textit{e.g.} $M_{i}, M_{j}$) at one time.
%

\subsubsection{Cross-modal feature-margin loss}
%
To minimize the gap between the image and music spaces, we introduce cross-modal feature-margin loss to control the largest distance between representations of the two modalities.
Suppose we obtain the image representation $f^I$ and music representation $f^M$ from their sub-networks.
In order to eliminate their difference, we penalize the image-music pairs whose distances are larger than $\alpha$:
\begin{equation}
	\begin{aligned}
		\label{met}
		L_{CFM}  = \sum_{i=1}^N [||f^{I_i} - f^{M_i}||_2-\alpha]_+,
	\end{aligned}%
\end{equation}
where $[\cdot]_+$ = max(0, $\cdot$). 
$\alpha$ is a threshold to manipulate the maximum tolerable distance.

\subsection{Single-modal Emotion Metric Learning}
\label{ssec:Single-modal}

Besides the cross-modal similarity relationship, we also enforce the embedding space to preserve the continuous emotion relationships for each modality when the VA labels are available. To achieve this goal, we minimize the feature-ratio loss within each modality. Differently, the distances between features from the same modality are computed based on their Euclidean distance between VA labels. The single-modal feature-ratio losses based on image and music representations are respectively definded as:
%
%
\begin{equation}
	\begin{aligned}
		\label{met}
	L_{SFR\_I} = \sum_{i=1}^n \left\{ \log \frac{D(f^{I_i}, f^{I_{j}})}{D(f^{I_i}, f^{I_{k}}) }-\log \frac{D(y^{I_i}, y^{I_{j}})}{D(y^{I_i}, y^{I_{k}}) } \right\}^2,
	\end{aligned}%
\end{equation}

\begin{equation}
	\begin{aligned}
		\label{met}
	L_{SFR\_M} = \sum_{i=1}^m \left\{ \log \frac{D(f^{M_i}, f^{M_{j}})}{D(f^{M_i}, f^{M_{k}}) }-\log \frac{D(y^{M_i}, y^{M_{j}})}{D(y^{M_i}, y^{M_{k}}) } \right\}^2,
	\end{aligned}%
\end{equation}
where $D(\cdot)$ means the squared Euclidean distance and $i \neq j \neq k$.
In the loss of each modality, an anchor and two neighbors take part in the loss computing.
By approximating the ratio between the distances of VA labels, the learned embedding space can reflect the emotion similarity of each modality.

\subsection{Embedded Multi-task Regression}
\label{ssec:Regression}

After learning the shared latent embedding space, we can jointly predict the matching similities and the concrete VA values.

\subsubsection{Cross-modal similarity MSE loss}
The similarity predictor is used to predict the matching similarity between a pair of image and music clip. This network is composed of three fully connected layers with BatchNorm, using Relu as the activation function except for the last layer which uses Sigmoid. Taking the concatenated image-music embeddings as input, the similarity predictor aims to minimize the following mean squared error (MSE) loss:
\begin{equation}
L_{Sim}=\frac{1}{N} \sum_{i=1}^N\left(S(I_i, M_i)-\hat{S}(I_i, M_i)\right)^{2},
\end{equation}
where $\hat{S}(I_i, M_i)$ is the predicted similarity of the $i^{th}$ image-music pair, while $S(I_i, M_i)$ is the corresponding ground truth, and $N$ is the total amount of matching pairs.

\subsubsection{Single-modal VA MSE loss}
The VA predictor is used to predict the VA values of images and music clicps. This part of the network is also composed of three fully connected layers with BatchNorm, using Relu as the activation function except for the last layer which uses Sigmoid. Taking the image or music embeddings as input, the predictor minizes a similar MSE loss:
\begin{equation}
L_{IVA}=\frac{1}{n} \sum_{j=1}^{n}\left(y^{I_i}-\hat{y}^{I_i}\right)^2,
L_{MVA}=\frac{1}{m} \sum_{j=1}^{m}\left(y^{M_j}-\hat{y}^{M_j}\right)^2,
\end{equation}
%
where $\hat{y}^{I_i}$ and $\hat{y}^{M_j}$ are the predicted VA values for image $I_i$ and music clip $M_j$.

\subsection{CDCML Optimization}
\label{ssec:Learning}
We can classify the loss functions mentioned above into two families, named similarity family $\mathcal{L}_{SF}$ and VA family $\mathcal{L}_{VAF}$.
If we only have similarity labels of image-music pairs, we can use $\mathcal{L}_{SF}$, which includes $L_{CFR}$, $L_{CFM}$, and $L_{Sim}$, to train our framework end-to-end for matching images and music clips.
If the VA labels of each modality are also available, we can simultaneously optimize both $\mathcal{L}_{SF}$ and $\mathcal{L}_{VAF}$, where $\mathcal{L}_{VAF}$ contains $L_{SFR\_I}$, $L_{SFR\_M}$, $L_{MVA}$, and $L_{IVA}$.
Therefore, with available similarity and VA labels, our CDCML framework can be optimized by minimizing the following total loss:
\begin{equation}
\mathcal{L}_{CDCML}= L_{CFR}+L_{CFM}+L_{Sim}+L_{SFR\_I}+L_{SFR\_M}+L_{MVA}+L_{IVA}.
\end{equation}
With the total loss, the embedding space and label space can be well optimized for the final prediction of multiple tasks.

\section{Experiments}
\label{sec:experiment}
In this section, we first introduce the experimental settings, including evaluation metrics, baselines, and implementation details, and then quantitatively compare the performance of the proposed cross-modal deep continuous metric learning (CDCML) method and several state-of-the-art approaches, followed by some ablation studies and visualization.

\subsection{Experimental Settings}

\subsubsection{Evaluation Metrics}
We employ mean squared error (MSE) and mean absolute error (MAE) to evaluate the effectiveness of the proposed CDCML method for image-music matching and VA prediction: $MSE=\displaystyle\frac{1}{t} \sum_{i=1}^{t}\left(l_{i}-\hat{l}_{i}\right)^{2}$, $MAE=\displaystyle\frac{1}{t} \sum_{i=1}^{t}\left|l_{i}-\hat{l}_{i}\right|$, where $\hat{l}_i$ represents the predicted value, $l_i$ is the ground truth label, and $t$ is the number of testing samples. MSE represents the sample standard deviation of the differences between predicted values and ground truth values. MAE is an arithmetic average of the absolute errors. Smaller MSE/MAE values represent better results.

%
%

\subsubsection{Baselines}
To compare CDCML with the state-of-the-art approaches for image and music matching, we select the following methods as baselines. \textbf{(1) SP-Net}, separately train two VA prediction models for image and music, calculate the corresponding Euclidean distance based on the predicted VA values for an image-music pair, and then obtain the matching similarity. Please note that SP-Net is trained only using VA labels. When the VA labels are unavailable, it does not work anymore. (2) \textbf{$L^3$-Net}~\cite{arandjelovic2017look} and (3) \textbf{ACP-Net}~\cite{verma2019learning}, extract features for image and music, fuse/concatenate the extracted features, and pass through several fully-connected (FC) layers to obtain the final similarity prediction. The differences lie in the input to the music feature extractors and the number of FC layers. Since $L^3$-Net and ACP-Net are initially designed for ``general audio-visual correspondence'' and ``affective audio-visual correspondence'' with 2-class output (\textit{i.e.} true or false correspondence), we replace the cross-entropy loss with MSE loss. Following ACP-Net~\cite{verma2019learning}, we feed the learned features of both images and music clips by $L^3$-Net and ACP-Net into another VA predictor to compare the performance of VA prediction.

\begin{table*}[!t]
\caption{Performance of the proposed CDCML and the stare-of-art approaches on IMEMNet for continuous emotion-based image and music matching. The best results are emphasized in bold.}
\begin{tabular}{c|cc|cccc|cccc}
\hline
                         & \multicolumn{2}{c|}{Similarity}                                                    & \multicolumn{4}{c|}{Image   emotion}                                                                                           & \multicolumn{4}{c}{Music   emotion}                                                                                          \\
\multirow{-2}{*}{Method} & MSE                                               & \multicolumn{1}{c|}{MAE}      & V   MSE                             & \multicolumn{1}{c}{V   MAE} & \multicolumn{1}{c}{A   MSE} & \multicolumn{1}{c|}{A   MAE} & V   MSE                             & \multicolumn{1}{c}{V   MAE} & \multicolumn{1}{c}{A   MSE} & \multicolumn{1}{c}{A   MAE} \\ \hline
SP-Net                   & 0.135          & 0.301          & 0.048          & 0.165          & 0.054          & 0.186          & 0.026          & 0.120          & 0.020          & 0.114          \\
$L^3$-Net~\cite{arandjelovic2017look}                   &   0.095          & 0.232          &     0.058          & 0.183          & 0.085          & 0.232      & 0.034          & 0.143          & 0.028          & 0.136          \\
ACP-Net~\cite{verma2019learning}                  & 0.086          & 0.222          &    0.062          & 0.195          & 0.091          & 0.241       & 0.027          & 0.130          & 0.022          & 0.131          \\
\textbf{CDCML (Ours)}                     & \textbf{0.067} & \textbf{0.210} & \textbf{0.044} & \textbf{0.157} & \textbf{0.050} & \textbf{0.175} & \textbf{0.024} & \textbf{0.118} & \textbf{0.015} & \textbf{0.099}            \\ \hline
\end{tabular}
\label{tab:SOA}
\end{table*}

\begin{table*}[!t]
	\centering
	\setlength\tabcolsep{5.5pt}
	\caption{Ablation studies of different components in CDCML on IMEMNet. `Sim', `VA', `CFR', `CFM', and `SFR' denote the cross-modal similarity MSE loss, single-modal VA MSE loss, cross-modal feature-ratio loss, cross-modal feature-margin loss, and single-modal feature-ratio loss, respectively. `$\surd$' means the corresponding loss is utilized in the training process.}
	\vspace{2mm}
	\begin{tabular}{c c c c c| c c | c  c c  c| c c c c }
		\toprule
		\multirow{2}*{Sim} & \multirow{2}*{VA}  & \multirow{2}*{CFR}  & \multirow{2}*{CFM} &\multirow{2}*{SFR} & \multicolumn{2}{c|}{Similarity} &\multicolumn{4}{c|}{Image emotion}& \multicolumn{4}{c}{Music emotion}       \\
		      &    &    & &    &  MSE   &   MAE  & V MSE   &V MAE   &A MSE    &A MAE   & V MSE   & V MAE   & A MSE    &A MAE      \\ \midrule
	$\surd$	&   &            &           &         &  0.083 & 0.239 & 0.060 & 0.195 & 0.087 & 0.239 & 0.039 & 0.163 & 0.046 & 0.173 \\
	$\surd$	&   &     $\surd$       &           &         &       0.074 & 0.231 & 0.058 & 0.187 & 0.075 & 0.225 & 0.034 & 0.153 & 0.042 & 0.163 \\
	$\surd$	&   &     $\surd$       &  $\surd$          &         & 0.072 & 0.227 & 0.057 & 0.186 & 0.074 & 0.229 & 0.034 & 0.153 & 0.042 & 0.162 \\ \midrule
	$\surd$ &    $\surd$  &       &    & & 0.080 & 0.233 & 0.046 & 0.158 & 0.052 & 0.180 & 0.026 & 0.120 & 0.017 & 0.104 \\
	$\surd$	&   $\surd$  &   $\surd$    &  $\surd$&  $\surd$ &  \textbf{0.067} & \textbf{0.210} & \textbf{0.044} & \textbf{0.157} & \textbf{0.050} & \textbf{0.175} & \textbf{0.024} & \textbf{0.118} & \textbf{0.015} & \textbf{0.099}            \\
		\bottomrule
	\end{tabular}
\label{tab:ablation}
\end{table*}

\subsubsection{Implementation Details}
As shown in Figure~\ref{fig:pip}, our model consists of two branches: the image branch and the music branch, which are used to respectively extract visual and audio features. After metric learning, the embeddings in the shared latent space are followed by two functional sub-networks: the similarity predictor and the VA predictor.

\textbf{The image branch} is based on Resnet-50. We drop the original classification layer and add one additional FC layer to obtain the final 512-dimensional visual features. Each image is resized to a predefined size of [224$\times$224$\times$3] before passing to Resnet-50.


\textbf{The Music Branch} is based on Resnet-18. We also drop the classification layer and add one FC layer to extract 512-dimensional audio features. Different from images, the input of the music branch is a batch of basic music features. We first extract the [193,87]-dimensional music features, which are composed of 40 MFCCs, 12 chroma features, 7 spectral contrast features, 6 tonal centroid features, and 128 features obtained from the mel spectrogram. And then we tile the music feature to form a feature matrix in the size of [193$\times$87$\times$3].

\textbf{The similarity predictor and  VA predictor} are both composed of 3 fully connected layers to respectively predict the similarity of an image-music pair and the concrete VA values of an image or music clip. Each FC layer is followed by a BatchNorm layer and an activate function layer with Relu, except for the last output layer which activation function is Sigmoid. Dropout rate is set to 0.5.


The weights of the feature extractors (\textit{i.e.} ResNet-50 and ResNet-18) are initialized from models trained on ImageNet. The network is implemented in PyTorch and trained with SGD optimizer using a batch size of 128 with initial learning rate 1e-3. The learning rate decreases with a decay of 0.1 for every 10 epochs.




\subsection{Comparison with the State-of-the-art}

The comparison of the proposed CDCML method and several state-of-the-art approaches on IMEMNet is shown in Table~\ref{tab:SOA}. From the results, we have the following observations:

(1) SP-Net performs the worst on cross-modal similarity prediction, but obtains much better results on VA prediction than $L^3$-Net~\cite{arandjelovic2017look} and ACP-Net~\cite{verma2019learning}. This is reasonable because SP-Net separately trains two VA prediction models for image and music. On one hand, with the VA labels as full supervision, SP-Net can learn discriminative representations for both image and music. On the other hand, without the similarity as supervision, it performs much worse than the methods that use similarity as supervision, \textit{i.e.} $L^3$-Net, ACP-Net, and the proposed CDCML.

(2) Similar to~\cite{verma2019learning}, our results also show that ACP-Net~\cite{verma2019learning} outperforms $L^3$-Net~\cite{arandjelovic2017look} on the matching and music VA prediction tasks. ACP-Net extracts various acoustic features, such as MFCC, chroma, and spectral contrast, while $L^3$-Net only uses log-spectrograms. Further, ACP-Net employs more FC layers to better learn the mapping between concatenated features and the similarity. However, $L^3$-Net performs better than ACP-Net on image VA prediction. This is because ACP-Net employs a pre-trained model to extract visual features, while the visual feature extractor in $L^3$-Net is trainable.

(3) CDCML obtains the best performance on both cross-modal matching and single-modal VA prediction. Specifically, compared to ACP-Net~\cite{verma2019learning}, CDCML achieves 22.1\% and 5.4\% relative performance improvements on MSE and MAE, while the relative gains over SP-Net on the valence and arousal of images and music measured by MSE are 8.3\%, 7.4\% and 7.7\%, 25.0\%, respectively. These results demonstrate the superiority of the proposed CDCML. The performance improvements benefit from the advantages of CDCML. First, it learns a shared latent embedding space which preserves the cross-similarity and single-modal emotion relationships in the label space. As a result, the embeddings are more discriminative for our task. Second, the embedded multi-task regression enables to learn a better similarity predictor and VA predictor with the joint supervision of similarity and VA labels.


\subsection{Ablation Studies}

We conduct in-depth ablation studies to systematically analyze the effectiveness of different components in CDCML.
The experimental results are shown in Table ~\ref{tab:ablation}.
If only cross-modal similarity labels between image and music are provided, we can train the network by optimizing cross-modal similarity MSE loss, cross-modal feature-ratio loss, and cross-modal feature-margin loss.
As shown in the first part of the table, cross-modal feature-ratio loss can notably improve the performance on similarity prediction (\textit{e.g.} 10.8\% relative gains on MSE).
Besides, what is pleasantly surprised is that the results of on VA prediction are also significanlty improved with the supervision of cross-modal feature-ratio loss in the shared embedding space.
It demonstrates that the loss makes the feature embeddings more discriminitive not only for cross-modal matching but also for single-modal VA prediction.
Note that cross-modal feature-margin loss is proposed to reduce the gap between different modalities by setting a maximum margin between features, so the performance of the output from shared FC layers is improved.

When VA labels are also provided, we can add several supervisions in both the embedding space and the VA space, as shown in the second part of Table ~\ref{tab:ablation}.
With the penalties of cross-modal similarity MSE loss and single-modal VA MSE loss on the final output, the embedded multi-task regression obtains better performance than that of using only one loss.
Apart from directly using VA label in single-modal VA MSE loss, we also use single-modal feature-ratio loss to manipulate the distance between features of the same modality based on the similarity of the VA labels.
It is obvious that the overall performance is further improved with the two losses that are based on VA labels, especially the results of VA prediction.
The effectiveness of single-modal feature-ratio loss indicates the importance of feature distribution in the latent embedding space.

\begin{figure*}[t]
	\centering
	\includegraphics[width=0.7\linewidth]{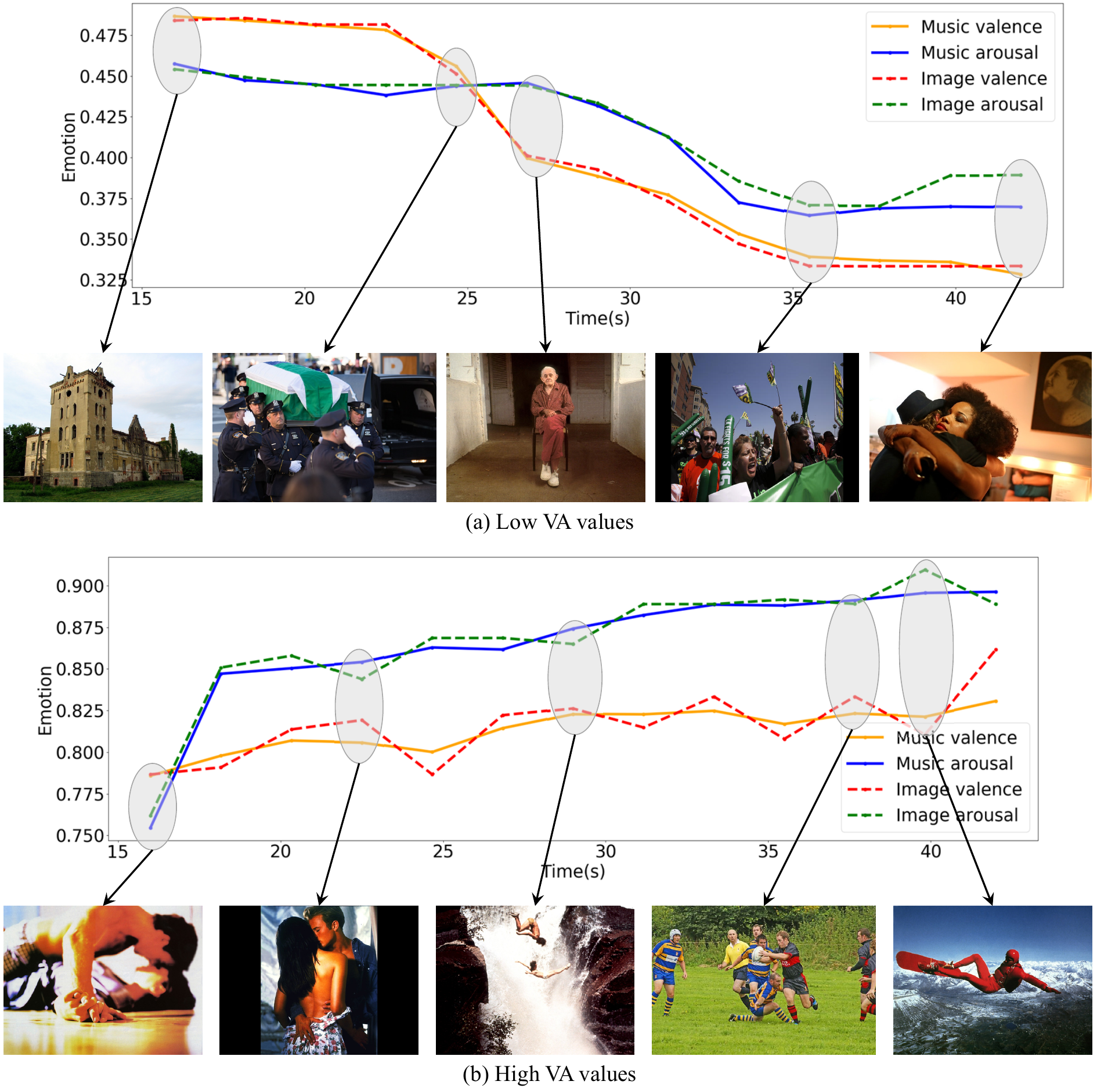}\\
	\vspace{2mm}
	\caption{Visualization of emotion-based image and music matching results in the VA space (best viewed in color). For each example, the upper part shows the emotion curves of both image and music over time. In the lower part, the images with high matching similarities to corresponding music clips are shown.}
	\label{fig:visualization}
\end{figure*}


\subsection{Visualization}
We vividly visualize the matching results between image and music based on continuous VA emotions in Figure~\ref{fig:visualization}.
We can observe that although the emotions of different music clips may change dramatically, the proposed CDCML method can well match suitable images for each music clip with similar emotions.

In Figure~\ref{fig:visualization} (a), the music's VA values are relatively low, indicating a negative emotion (\textit{e.g.} sadness). It is clear that the matched images also have similar emotions. For example, the funeral and lonely elder lady both make people feel sad. In Figure~\ref{fig:visualization} (b), the music's emotions are represented with high VA values, corresponding to a positive emotion (\textit{e.g.} excitement). Meanwhile, the matched images tend to be passionate, such as the kisses and extreme sports, which can easily evoke exciting emotions. The qualitative
matching results further demonstrate the effectiveness of the proposed CDCML method for matching image and music based on emotions.






\section{Conclusion}
\label{sec:conclusion}
In this paper, we aimed to study continuous emotion-based image and music matching in an end-to-end manner. To learn a shared latent embedding space, we proposed cross-modal deep continuous metric learning (CDCML) by preserving the cross-modal similarity and single-modal emotion relationships. The embedded multi-task regression can simultaneously predict the matching similarity and VA values. To evaluate the effectiveness, we constructed a large-scale dataset, termed IMEMNet. The extensive experiments on IMEMDnet demonstrate that CDCML achieves 22.1\% and 5.4\% relative performance improvements on MSE and MAE for matching similarity prediction as compared to the best state-of-the-art method (\textit{i.e.} ACP-Net~\cite{verma2019learning}). In future studies, we plan to model the sequential information of different music clips in a whole song using LSTM-based techniques. In addition, we will study a more practical image and music matching based on both emotions and semantics. How to deal with incremental training image-music pairs is also worth exploring.

\begin{acks}
This work is supported by the National Natural Science Foundation of China (Nos. 61701273, 61876094, U1933114), Berkeley DeepDrive, the Major Project for New Generation of AI Grant (No. 2018AAA0100403), Natural Science Foundation of Tianjin, China (Nos. 18JCYBJC15400, 18ZXZNGX00110), the Open Project Program of the National Laboratory of Pattern Recognition (NLPR), and the Fundamental Research Funds for the Central Universities.
\end{acks}

\balance\bibliographystyle{ACM-Reference-Format}
\bibliography{sample-base}


\end{document}